\documentclass{sigkddExp}

\usepackage{soul}
\usepackage{url}
\usepackage[utf8]{inputenc}
\usepackage[small]{caption}
\usepackage{graphicx}
\usepackage{amsmath}
\usepackage{booktabs}
\usepackage{booktabs} 
\usepackage{graphicx}
\usepackage{subfigure}
\usepackage{booktabs}
\usepackage{threeparttable}
\usepackage{bm}
\usepackage{multirow}
\usepackage{makecell}
\usepackage{diagbox}
\usepackage{color}

\begin{document}
%

\title{Surf or sleep? Understanding the influence of bedtime patterns on campus}
%

\numberofauthors{2}
%


\author{
%
\alignauthor Teng Guo, Linhong Li, Dongyu Zhang \\
       \affaddr{School of Software,}\\
       \affaddr{Dalian University of Technology,}\\
       \affaddr{Dalian 116620, China}\\
       \email{teng.guo@outlook.com; li.linhong@outlook.com; zhangdongyu@dlut.edu.cn}
\alignauthor Feng Xia\thanks{Corresponding author} \\
       \affaddr{School of Engineering, IT and Physical Sciences,}\\
       \affaddr{Federation University Australia,}\\
       \affaddr{Ballarat, VIC 3353, Australia}\\
       \email{f.xia@acm.org}
}
\date{30 July 1999}
\maketitle
\begin{abstract}
Poor sleep habits may cause serious problems of mind and body, and it is a commonly observed issue for college students due to study workload as well as peer and social influence.
Understanding its impact and identifying students with poor sleep habits matters a lot in educational management.
Most of the current research is either based on self-reports and questionnaires, suffering from small sample size and social desirability bias, or the methods used are not suitable for the education system.
In this paper, we develop a general data-driven method for identifying students' sleep patterns according to their Internet access pattern stored in the education management system and explore its influence from various aspects.
First, we design a Possion-based probabilistic mixture model to cluster students according to the distribution of bedtime and identify students who are used to stay up late.
Second, we profile students from five aspects (including eight dimensions) based on campus-behavior data and build Bayesian networks to explore the relationship between behavioral characteristics and sleeping habits.
Finally, we test the predictability of sleeping habits.
This paper not only contributes to the understanding of student sleep from a cognitive and behavioral perspective but also presents a new approach that provides an effective framework for various educational institutions to detect the sleeping patterns of students.
\end{abstract}

\section{INTRODUCTION}
Sleep matters a lot for health and well-being \cite{simon2018sleep,zhao2019association,althoff2017harnessing}.
Poor sleep habits like insomnia can easily cause serious problems like worsened health-related quality of life and poor health outcomes, especially for college students with psychological immaturity \cite{combs2016insomnia}.
Research have shown that the majority of college students suffer from sleep disorders like feeling tired, fatigue, or daytime sleepiness and sleeping less than eight hours per night during the school week \cite{Becerra2018Sleepless,kelley2015synchronizing}.
68\% of American teenagers report that their sleep time each night is below the recommended eight-ten hours \cite{cousins2019does} and in East Asia, the sleep time of adolescents on weekdays night is below six hours \cite{lo2016cognitive,cousins2018memory}.
Therefore, a significant topic of educational research is identifying college students with bad sleep habits timely \cite{tavernier2014sleep,alimoradi2019internet}, and understanding its influence.

However, detecting sleep patterns accurately faces tremendous challenges.
Previous research are mainly based on questionnaires, which is time-consuming and costly making, and it is hard to be scale to a lot of students.
In the last decade, scientists propose a variety of techniques for sleep tracking, and some related products are already in use \cite{al-salman2019detecting,okano2019sleep}. These techniques can only be used for small-scale analysis due to the high cost of the associated equipment.
Some studies have tried to do large-scale sleep-related research through a data-driven approach like \cite{althoff2017harnessing}.
However, such data is difficult to access except for large companies like Microsoft, Google, etc.
A data-driven framework for sleep habit detection and analysis that can serve the daily management of universities is urgently needed.

During the past decade, we have witnessed an increased utilization of electronic devices like smartphones and tablets \cite{bedru2020big}.
As they become more lightweight, people may easily use these devices even in bed.
Electronic media use in bed after bedtime is more common in younger compared to older age groups \cite{bjorvatn2017age}.
Research show that more than 90\% of young people are used to playing mobile phones or tablets before bedtime \cite{gronli2016reading,fossum2014association}.
Meanwhile, for college students, the campus network is a good choice because of its low price and faster internet speed, which generates tons of data of internet access logs stored in the local education management system.
This provides university management divisions a new way to detect the sleep patterns of college students.
However, new challenges are introduced as well.
First, these data are not easy to obtain due to privacy protection \cite{guo2020graduate,peters2020responsible}.
Second, the volume of data collected in this way is much larger than the traditional method.
Targeted methods need to be designed for efficient data mining.

In this paper, we are devoted to identifying students with the habit of staying up late and explore the influence of this habit.
First, we use the Internet access pattern to identify the bedtime of each student.
Second, we design a Possion-based probabilistic mixture model to cluster students according to the distribution of bedtime and identify students who stay up late.
Third, besides sleep habit, we profile students from five aspects in eight dimensions, including interest (reading status, app preference status, and surfing length status), orderliness (breakfast orderliness status and bath orderliness status), finance (financial status), academic performance (academic performance status) and gender.
Then we build Bayesian networks to explore the relationship between these characteristics and sleeping habits.
Finally, we design an experiment of predictive inference to estimate the predictability of sleep status.

Our contributions could be summarized as follows:
\begin{itemize}
\item We design a Possion-based probabilistic mixture model suitable for education management divisions to identify students' sleep patterns.
\item We profile students behavior from various aspects and build Bayesian networks to explore the relationship between these characteristics and sleeping habits.
\item We conduct comprehensive experiments on a large-scale educational dataset and the extensive results demonstrate the effectiveness.
\end{itemize}

This paper is organized as follows. In the next section, related work is reviewed. The methods are presented in Section 3. In Section 4, we introduce the experiment setting and analyze the results. In the final section, we present the discussion and conclusion of our work.

\section{Related work}
The sleep patterns of college students have attracted the attention of scholars from various fields. As the negative impact of poor sleep quality has been demonstrated, the main concern of scientists is the factors that affect sleep quality.
Buboltz et al. \cite{buboltz2001sleep} carry out a study focused on the sleep disturbance of college students. The results show that most college students suffer from sleep disturbance and women exhibit more sleep disturbance than man do. Moreover, they give some suggestions to reduce the negative influence on academic performance.
Tsai et al. \cite{tsai2004sleep} also focus on how gender can impact students' sleep.
Brown et al. \cite{brown2002relationship} use regression modeling to capture the relationship between sleep hygiene and sleep practices. The results demonstrate that variable sleep schedules, going to bed thirsty, environmental noise, and worrying can all impact the quality of sleep.
Tavernier et al. \cite{tavernier2015longitudinal} explore how the social tie can impact sleep and the results show that positive social ties can contribute to good sleep quality.
Patrick et al. \cite{patrick2018energy} focus on the influence of energy drinks and binge drinks on sleep quality. The results show that drinking the above two drinks can easily lead to poor sleep quality and fatigue the next day.
Meanwhile, Schneider et al. \cite{schneider2020adolescent} focus on adolescents and explore the relationship between sleep characteristics and body-mass index. In this study, they emphasize the importance of gender in related research.
Orzech et al. \cite{orzech2016digital} proves that longer digital media use two hours prior to bedtime leads to reduced sleep and later bedtime. They obtain an interesting conclusion that more diverse media uses are associated with longer sleep durations.
Moreover, some scholars have focused on sleep patterns and their impact on daily life. Kelley et al. \cite{kelley2015synchronizing} explain adolescents' fatigue and irritability by the sleep problems caused by the disorder of their biological time and social time.
Althoff et al. \cite{althoff2017harnessing} carry out a convincing experiment based on a search engine Bing from Microsoft including 3 million nights of sleep and 75 million interaction tasks. The results show how sleep deprivation can seriously affect daily performance. However, as mentioned before, such type of data is inaccessible for education management divisions.

Thanks to the development of technologies, more and more electronic devices are appearing in our daily lives \cite{burr2020digital} and scientists try to use these advanced technologies (e.g. remote bio-signal monitoring technology) to solve sleep problems.
Vroon et al. \cite{vroon2017snoozle} propose an actuated robotic pillow to improve the quality of sleep through enhancing the interaction with users.
De Arriba et al. \cite{de2016calculation} propose a smartphone-based indicator that can calculate sleep patterns automatically. They try to use this indicator to facilitate the construction of software services in order to improve the efficiency of the learning process.
Liu et al. \cite{liu2017development} propose a mobile sleep-management system integrated with self-regulated learning strategies and cognitive behavioral therapy. This system can help students to improve their sleep quality by improving their daily schedule and modify strategies to cultivate good learning and healthy lifestyle habits.
Zhao et al. \cite{zhao2020design} design a smart-home system based on the Internet of things, which realizes the function that judges the user's mood and automatically plays corresponding music through facial expression recognition.
However, the methods mentioned above are difficult to be used for large-scale sleep detection on campus due to their high cost.

\section{Methods}
In this section, we describe the models used in the experiment. First, we introduce a mixed probability model used to cluster students' sleep habits. Then, we introduce the Bayesian network model used to analyze the influence of different sleeping habits.
\subsection{Sleep Pattern Recognition}
\subsubsection{Aggregated Sleep Count}
\begin{figure}
	\subfigure[Bedtime of a student in a month: this is the calendar of a certain month and the number in the middle of the square is the bedtime of the day.]{
		\begin{minipage}{7cm}
			\centering
			\includegraphics[height=3.5cm]{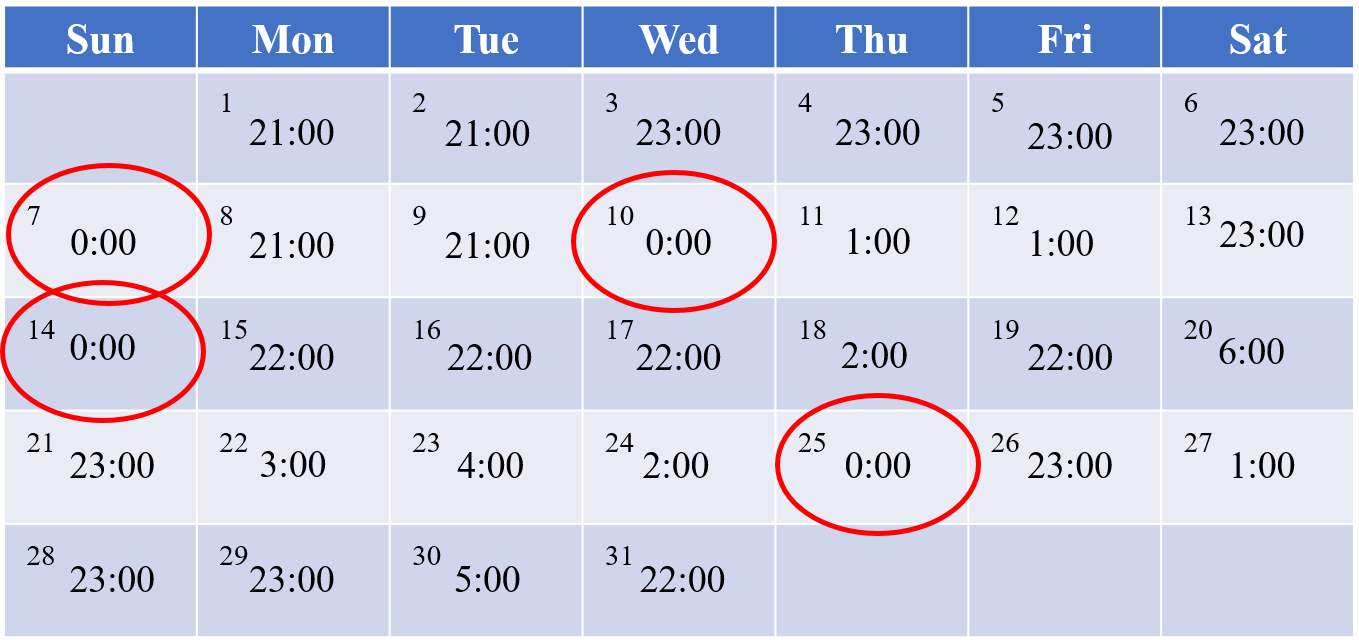}
        \label{rili}
		\end{minipage}
	}
	\subfigure[Aggregated Sleep Count of a student in a month: this is the aggregated sleep count of each time period during the month.]{
		\begin{minipage}{7cm}
			\centering
			\includegraphics[height=4.5cm]{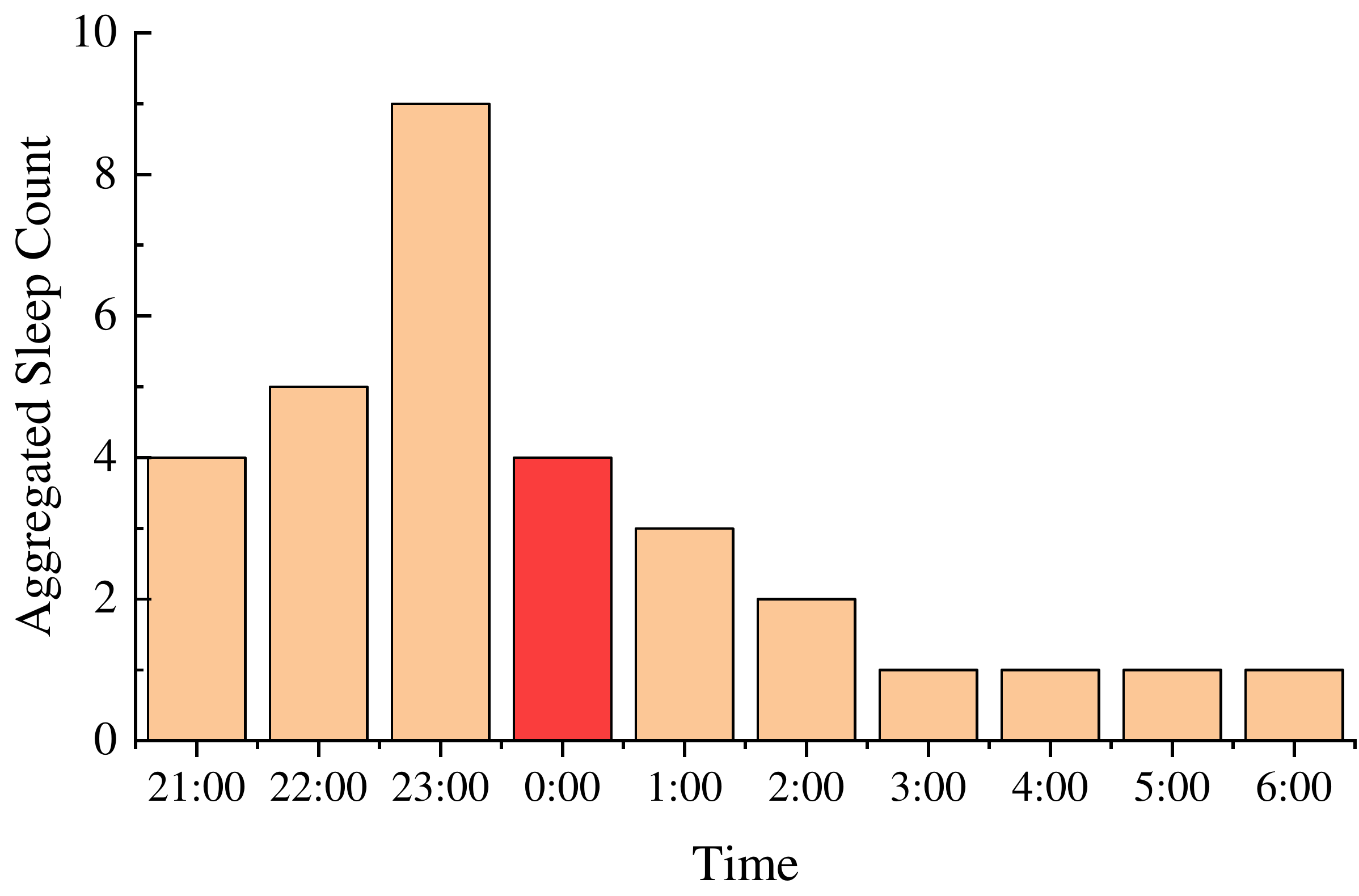}
        \label{fenbu}
		\end{minipage}
	}	
	\caption{Illustration for aggregated sleep count.}
	\label{asc}
\end{figure}

For universities that have campus-dedicated networks deployed, web logs can be used to calculate the bedtime of students every night based on the assumption that students will access the Internet through mobile phones or computers before bedtime \cite{bartel2019altering,gronli2016reading,fossum2014association}. In other words, we can use the time of the last network signal every day to quantify the time students go to bed.
To mine sleep patterns of different students, we define $aggregated$ $sleep$ $count$, which represents the total number of days a student goes to sleep at a particular time slot.
For example, in a certain month, daily bedtime is shown in Figure \ref{rili}.
Let's take 0:00 as an example. In this month, there are 4 days that sleep at 0:00 o'clock shown in Figure \ref{rili} with the red circle, the aggregated sleep count of 0:00 in this month is 4 shown in Figure \ref{fenbu} with a red bar.
The $aggregated$ $sleep$ $count$ of each time period is shown in Figure \ref{fenbu}.

\subsubsection{Mixture Model with Gamma Priors}
Inspired by \cite{park2018understanding}, we design a Poisson mixture model to mine the pattern on their aggregated sleep counts in each time period.
For student $i$, we let $s_i$ be the vector of aggregated sleep counts, where $ i = 1,2,3...N$. The dimensionality $D$ of each vector is the number of time windows ($D = 16$ in this case that we divided the time into 16 segments from 9:00 pm to 4:30 am and the time span of each segment is 30 minutes because: 1, the last class generally is over at 9:00 pm. 2, the sun rises at around 4:30 am).
Thus, our data consists of $N$ students each with $D$-dimensional vectors of aggregated sleep counts.

To mine the pattern behind this data, we build a probabilistic mixture model with Poisson components.
In terms of notation, we let $M$ be the number of components with an index $m = 1,2,3...M$.
The unobserved latent variable $z_i$ takes values from the set \{$1,2,...M$\} that corresponds to the latent component of student $i$.
Each component consists of a vector of Poisson rate parameters,
$\bm{\lambda_m} = [\lambda_{m1},.....\lambda_{md},.....\lambda_{m16}],$ where $d$ from $[1,2,3.....D]$ represents a time window mentioned before.
For example, a low value for $\lambda_{md}$ represents that students rarely sleep during time window $d$. On the contrary, a high value for $\lambda_{md}$ represents that students always sleep during time window $d$.

In this study, the Bayesian approach is used to fit our mixture model.
To encourage the model to avoid degenerate solutions, we use Gamma prior distribution for the rate parameter $\lambda_{md}$ according to the conjugate prior theory.
More precisely, we use hyper parameters $\alpha=1.1$ and $\beta=0.1$ for the Gamma distribution, which makes it a step-like function that puts zero probability mass at $\lambda_{md}$ and provides a relatively flat uninformative prior distribution over positive rate parameter values \cite{park2018understanding}.

According to the total probability formula, the probability of each student $i$ under this mixture model is:
\begin{equation}
\begin{aligned}
p(\bm{s_i}|\bm{\lambda}) = \sum_{m=1}^{M}p(\bm{s_i}|z_i=m,\bm{\lambda_m})p(z_i=m)
\end{aligned}
\end{equation}
where $p(z_i = m)$ is the marginal mixing weight for each component. Assuming conditional independence of aggregated sleep count $s_{md}$ given component $m$,  the probability distribution of each component can be decomposed as:

\begin{equation}
\begin{aligned}
p(\bm{s_i}|z_i=m,\bm{\lambda_m}) = \prod_{d=1}^{D}p(s_{md}|\lambda_{md}, z_i=m)
\end{aligned}
\end{equation}
\subsubsection{Parameter Estimation}
Expectation-Maximization (EM) algorithm, which is widely used in fitting a mixture model to data\cite{dempster1977maximum,mclachlan2007algorithm}, is used in this experiment to estimate the parameter $\bm\lambda_m$.
More specifically, the optimization objective in this experiment is to maximize the product of the data likelihood times the prior probability.
This optimization process produces two results.
The first one is maximum a posteriori parameter estimates for the Poisson components in the model and the second one is membership weights $\omega_{im}$ that is the probability that student $i$ belongs to component $m$.
Each iteration of the EM algorithm consists of two processes, the E (expectation) process and the M (maximization) process. In the E process, the probability of membership is computed for each component $m=1,2,...M$, for each student $i = 1,2,...N$.

\begin{equation}
\centering
\begin{aligned}
\omega_{im} &= p(z_i=m|\bm{s_i},\bm{\lambda},\alpha,\beta) \\
&\propto p(z_i=m,\bm{s_i},\bm{\lambda_m}|\alpha,\beta) \\
 &     \propto p(\bm{s_i}|z_i=m,\bm{\lambda_m})p(\bm{\lambda_m}|\alpha,\beta)p(z_i=m)
\end{aligned}
\end{equation}
where $\omega_{im}$ is the probability that student $i$ belongs to component $m$.
In the M process, conditioned on the set of membership probabilities $\omega_{im}$, we estimate each parameter via MAP (Maximum Posteriori) estimation:

\begin{equation}
\begin{aligned}
\bm{\hat\lambda_m} = \frac{\sum_{i}\omega_{im}(\bm{s_i}+\alpha - 1)}{\sum_{i}\omega_{im}(\beta+1)}
\end{aligned}
\end{equation}

\begin{equation}
\begin{aligned}
\hat{p}(z_i=m) = \frac{\sum_{i}^{N}\omega_{im}}{N}
\end{aligned}
\end{equation}

The outputs of the M process provide the input for the next E process, and thus, the cycle of E and M processes continue iteratively.

\subsection{Bayesian Network}
In this study, Bayesian network (BN) is used to explore the relevance between bedtime patterns and behavioral characteristics \cite{jensen1996introduction,friedman1997bayesian}.
Based on Markov property, the BN uses a set of local distributions to expresses the global joint distribution \cite{agrahari2018applications,chien2002using}.

\subsubsection{Bayesian network}
A BN consists of two parts: a directed acyclic graph (DAG) and a set of conditional probability distribution, which can be represented as $\mathcal{B} = \langle G_d,\bm{\Theta} \rangle$. The graph $G_d$ represents a directed acyclic graph whose vertices correspond to random variables $X_1,X_2,...X_n$ and whose edges represent direct dependencies between the variables, where $n$ represent the number of variable in $\mathcal{B}$. The $\bm{\Theta} = (\bm{\theta_i})_{1\leq i \leq n}  $ denotes the set of all parameters.

The nations we used are introduced as follows:
\begin{itemize}
\item $n$ represent the number of variable in $\mathcal{B}$ and each node $X_i$ takes $r$ possible values $x_1,x_2,...x_r$
\item $\pi(X_i)$, denoted the parent nodes of each variable $X_i$ in $G_d$, and has $q_i = \prod_{m \in \Omega_i}r_m$ possible configurations, where $\Omega_i=\{m;X_m \in \pi(X_i)\}$
\item $ch(X_i)$ denotes the children nodes of each variable $X_i$ in $G_d$
\item The conditional distribution of $ X_i|\pi(X_i)$ is defined by the matrix of probabilities $\bm{\theta_i} = (\theta_{ijk})_{1 \leq j \leq q_i, k \in (x_1,...x_r) }$:
\begin{equation}
\begin{aligned}
\theta_{ijk} = P(X_i = k|\pi(X_i)=j)
\end{aligned}
\end{equation}

\end{itemize}

For each $X_i$, $\bm{\theta}_{ij} = (\theta_{ijx_1},\theta_{ijx_2},...\theta_{ijx_r})$ is the vector of conditional probabilities defining the distribution of $X_i|\pi(X_i)=j$, where $\Sigma_{k=1}^{r} \theta_{ijk} = 1 $. The joint probability distribution of  $(X_1,X_2,...X_i)$ is given by:

\begin{equation}
\begin{aligned}
P(X_1,X_2,...X_i) = \prod_{i}^{d} P(X_i|\pi(X_i))
\end{aligned}
\end{equation}

\subsubsection{Structure Design}

\begin{figure}[b]
	\centering
	\includegraphics[width=6cm]{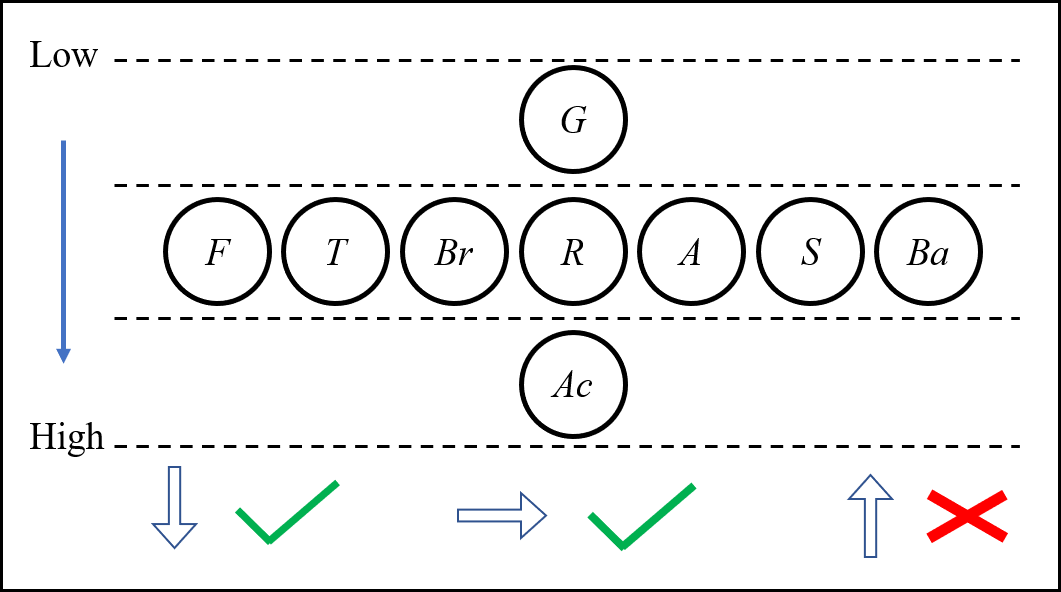}
	\caption{Three-layers raw structure for BN.}
	\label{raw}
\end{figure}

To explore the main problem proposed in this paper mentioned above, we first define nine binary variables used in our BN shown in Table \ref{variable} (Details are introduced in the following section).
We use student data for one semester in this paper.
Based on the causal Markov assumption \cite{Spirtes1993Causation}, we develop a three-layers raw structure, shown in Figure \ref{raw}, with two rules: (1), the directed edge can start from one node to another node on the same layer. (2), the directed edge can only point from the lower-layer to the upper level. In other words, we blacklist all outgoing edges from the upper-level nodes to lower level nodes. By construction, no node is allowed to be the parents of gender, and no node is allowed to be the child of academic performance. Such design is consistent with common sense.
\begin{table*}
\centering
\caption{Variables definition.}
\begin{tabular}{c|c|c}
\hline
Variable& Definition& Description \\
\hline
$G$ & Gender & Male or Female  \\
$R$ & Reading status & Classify students according to the number of books borrowed \\
$A$ & App preference status &\makecell[c]{Define online preferences based on usage time of\\ each app (Game app or Video app)} \\
$T$ & Surfing length status &Classify students according to their total length of surfing time\\
$Br$ & Breakfast Orderliness status& Classify students according to the number of times they eat breakfast\\
$Ba$ & Bath Orderliness status & Classify students according to the regularity of bathing\\
$F$ & Financial status &Classify students based on daily average spending\\
$Ac$ & Academic performance status & Classify students based on Grade-Point Average\\
$S$ & Sleep status & Classify students according to whether they stay up late\\
\hline
\end{tabular}
\label{variable}
\end{table*}

\subsubsection{Structure Learning}

In this research, we use BDeu (Bayesian Dirichlet equivalent uniform) score to evaluate the BN and use the hill-climbing strategy to optimize this score \cite{heckerman1995learning}. BDeu score is a variant of the BDeu score, which aims at maximizing the posterior probability of the DAG. We use MLE (maximum likelihood estimation) to fit our BNs \cite{article} in this research.

\section{Experiments and Results}
\subsection{Dataset}
The dataset used in this research includes 5,200 students from a Chinese university including freshmen students, sophomores, and juniors and contains a total of about 50 million records of data related to various behavior.
For the privacy concern, the students are already pseudonymous in the raw data. Removing the student with incomplete data, we have 4,249 students left classified into three groups: freshman (1,423 students), sophomore (1,580 students), and junior (1,246 students). This dataset consists of five types of data and the details are described as follows:

\subsubsection{Internet Access Data} For most students, the campus network is a good choice because of its low price and faster internet speed.
The related logs accurately record the student's online information. We use this data to profile students' surfing behavior. The experiment time is from November 2018 to April 2019 including 4,541,247 records data. To ensure the validity of the experimental results, we remove students with fewer related records. In this research, surfing length status $T$, app preference status $A$, and sleep status $S$ are calculated based on this data.

\subsubsection{Academic Performance Data} Students' academic performance is generally be recorded, which contains the grade,
credit of each course for each student. The academic performance data includes 1,048,575 records. In this research, we select the most recent grade data from the period of Internet access data to obtain academic performance data $Ac$.
\subsubsection{Demographic Data} For all schools, students are required to submit personal information at the time of admission, like hometown, gender, and nation. In this research, we take gender $G$ as a feature.

\subsubsection{Financial Data} Campus smart card, in most universities, is used as a recognition tool for his identification. Generally, smart cards can be used for any scene of consumption on campus and thus record tons of data for student consumption behavior, such as bathing and eating. Financial data include 22,176,513 records. In this research, breakfast orderliness status $Br$, bath orderliness status $Ba$, and financial status $F$ are obtained based on this data.

\subsubsection{Reading Data} Generally, every university has its own library that is the main source of book reading by students. Reading data include 17,209,329 records. Reading status $R$ in this research is obtained from this data.
\subsection{Student Clustering}
Below, the results of fitting a two-component Poisson mixture model are presented and discussed. Models with more components are explored in this experiment, but the two-components model mainly captures the primary modes of student's sleep patterns and more components models tend to split students into further subgroups without any significant additional insight.

\begin{figure*}
	\subfigure[Freshman]{
		\begin{minipage}{8.5cm}
			\centering
			\includegraphics[height=5.5cm]{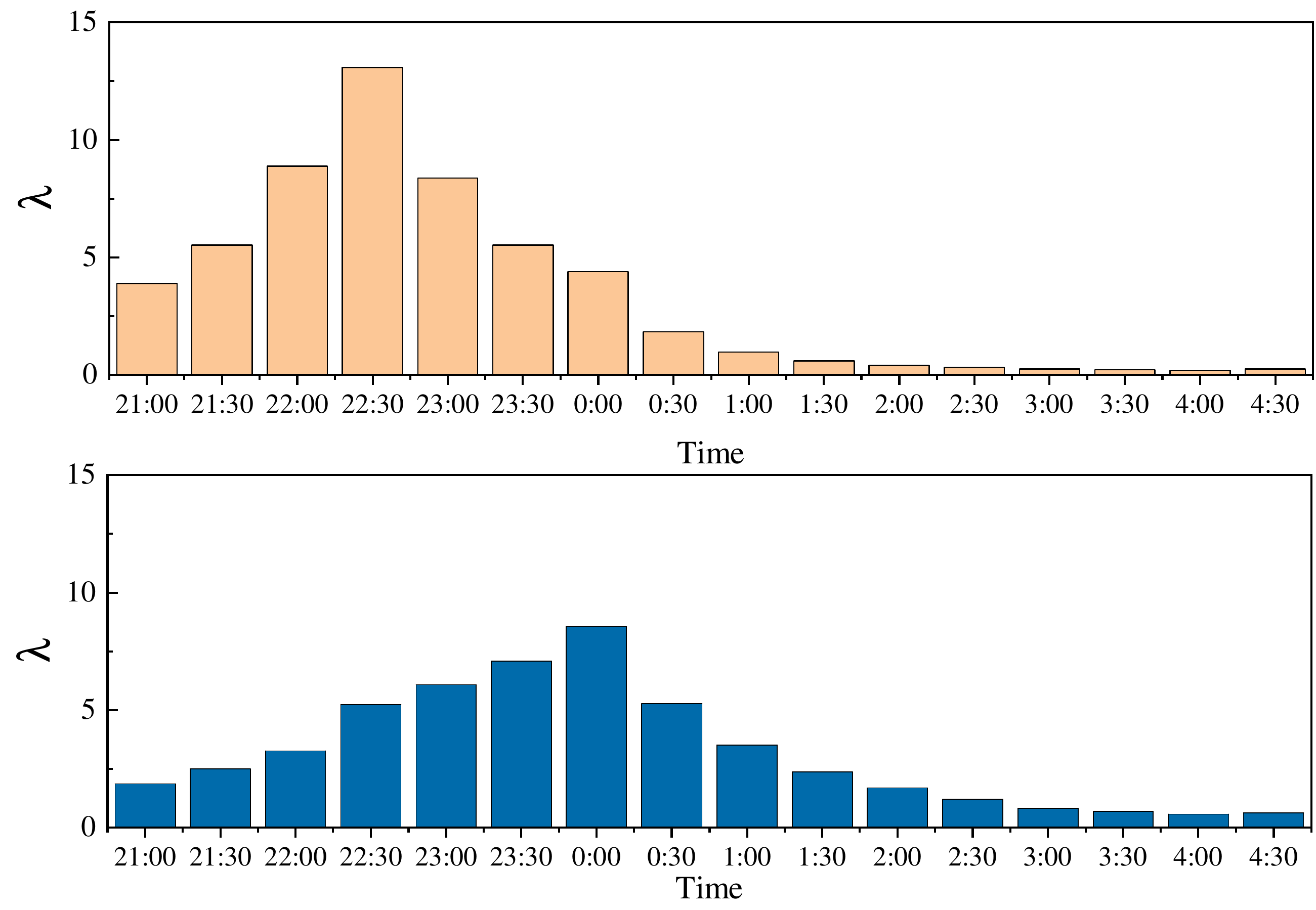}
		\end{minipage}
	}
        \subfigure[Sophomore]{
		\begin{minipage}{8.5cm}
			\centering
			\includegraphics[height=5.5cm]{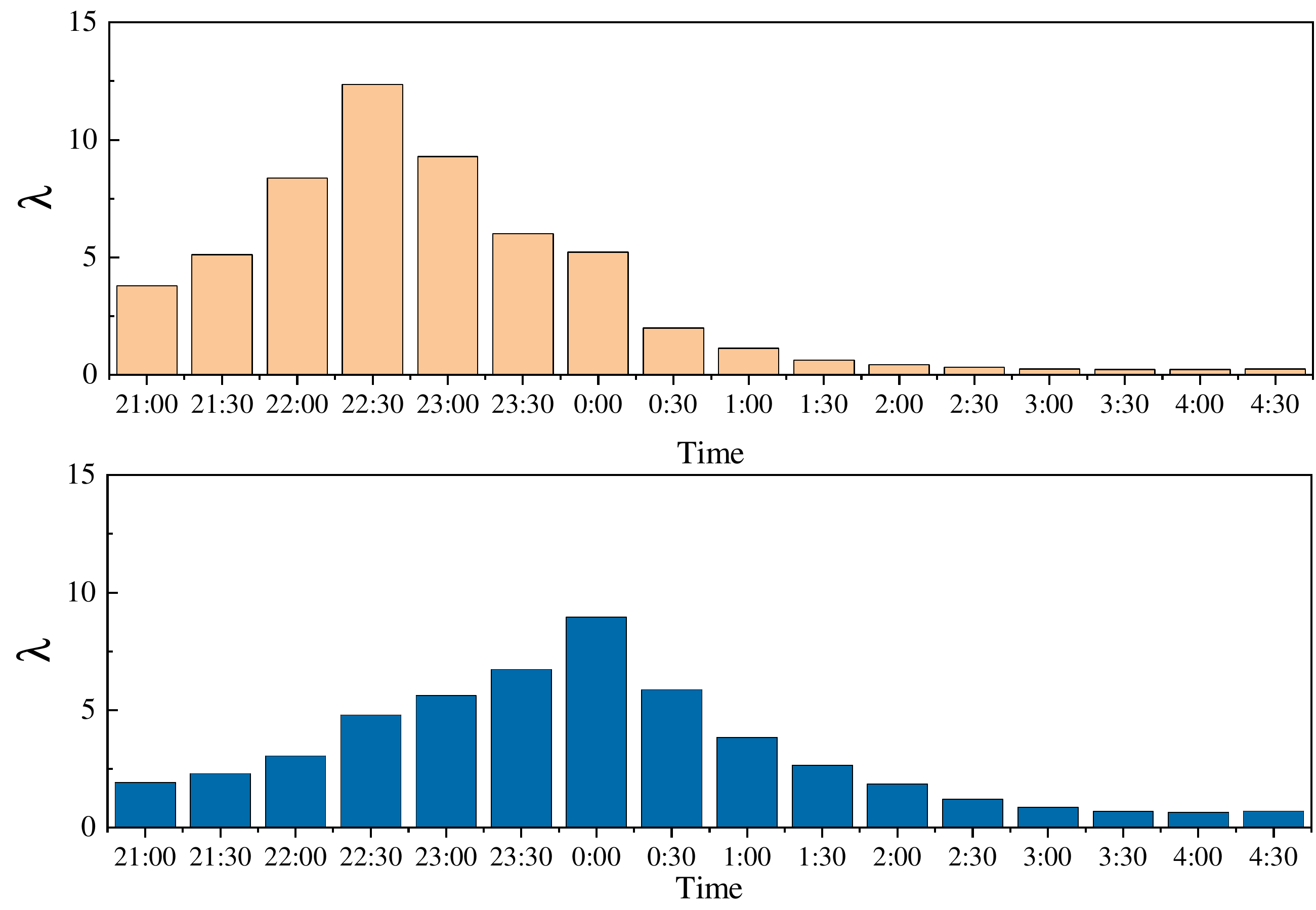}
		\end{minipage}
	}	
	\subfigure[Junior]{
		\begin{minipage}{8.5cm}
			\centering
			\includegraphics[height=5.5cm]{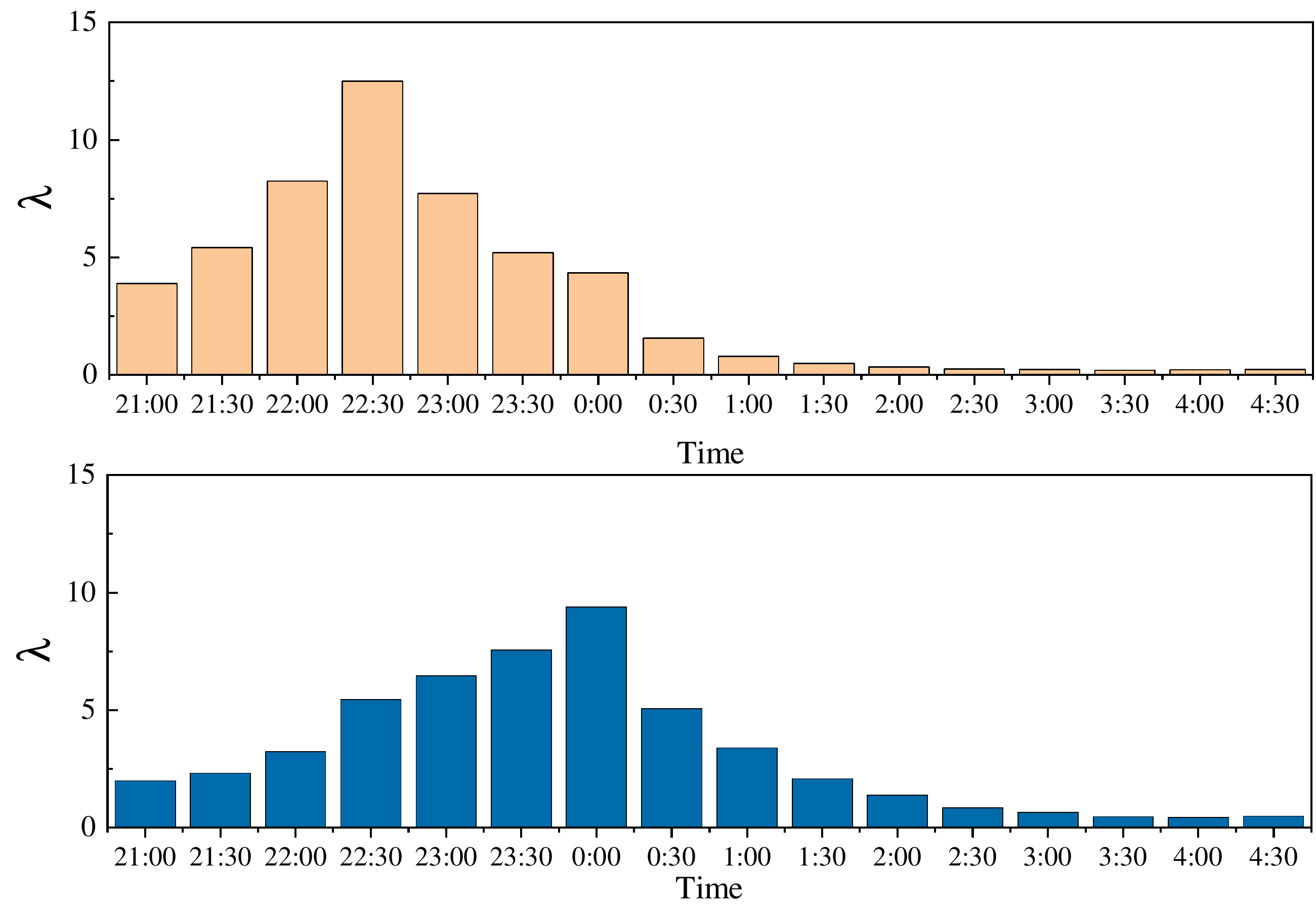}
		\end{minipage}
	}
	\subfigure[Total]{
		\begin{minipage}{8.5cm}
			\centering
			\includegraphics[height=5.5cm]{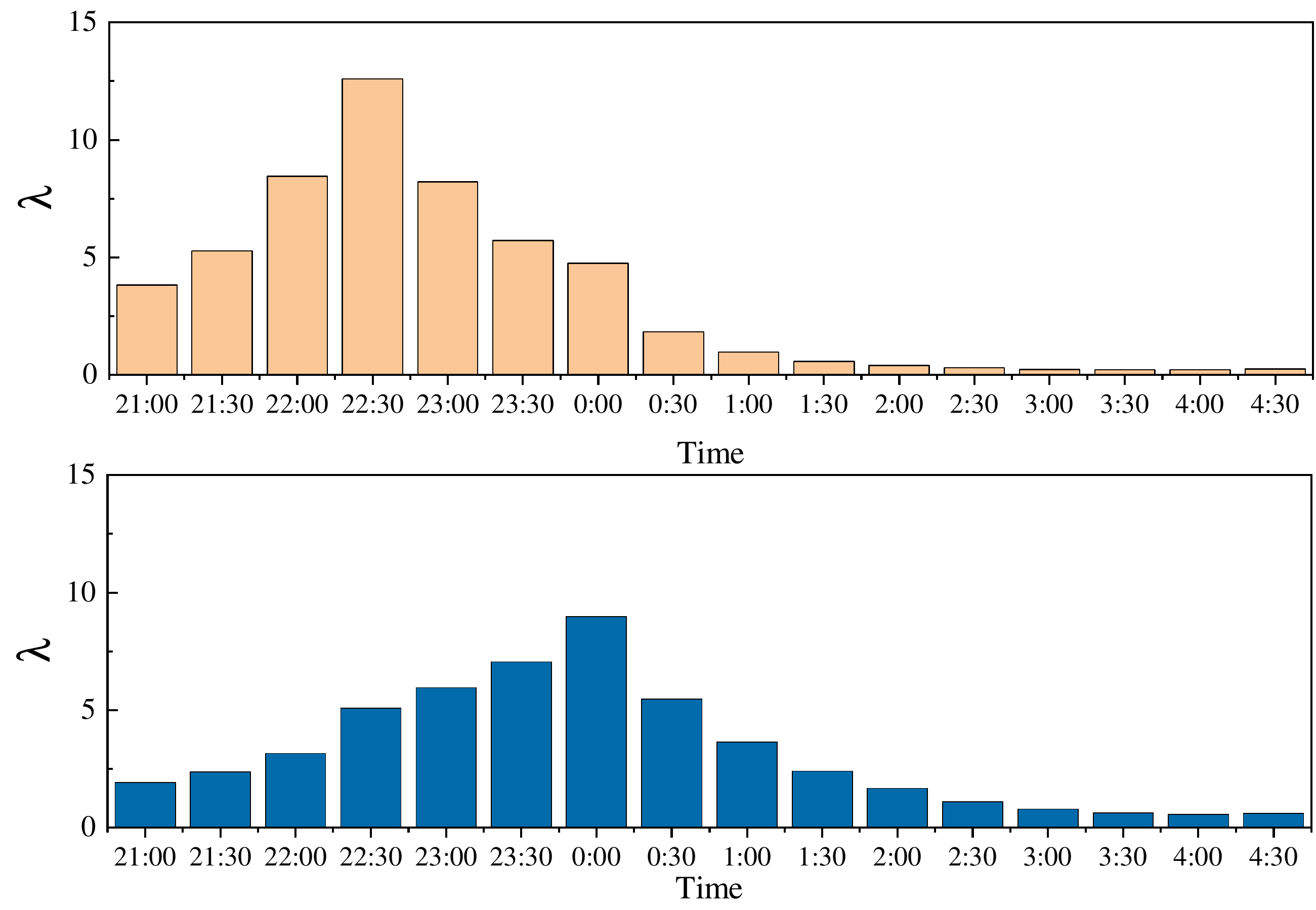}
		\end{minipage}
	}

	\caption{Clustering results of Poisson mixture model with two components.}
	\label{lambda}
\end{figure*}

\begin{table}
\centering
\caption{The number of student in each cluster.}
\begin{tabular}{cccccc}
\hline
 & stay-up  & non-stay-up & Total \\ \hline
    Junior  & 664 & 759  & 1423  \\
    Sophomore  & 807 & 773  & 1580    \\
    Freshman  & 714  & 532 & 1246    \\
    Total  & 2185 & 2064 & 4249   \\

\hline
\end{tabular}
\label{student number}
\end{table}

We set the threshold of component weight at 0.5 to classify student $i = 1,2,...N$ into one of the two components, i.e., if $\omega_{i1} \geq 0.5$, then student $i$ is classified into the group of staying up (where $m=1$ corresponds to the staying up group).
We first experiment individually for freshmen, sophomores, and juniors, and then experiment with all students together.
Figure \ref{lambda} shows the $\lambda$ of each component and Table \ref{student number} show the corresponding number of each components.
Two distinct sleep patterns can be seen for each group, as the $\lambda$ of Possion distribution represents the number of times things happen per unit of time.
Each pattern is an approximate bell curve. They can be distinguished according to two important characteristics shown as follows:
\begin{itemize}
\item First, one of the mixture components has a peak at 10:30 pm (shown as the yellow bar chart) and the other one has a peak at 0:00 am (shown as the blue chart).
\item Second, the mixture component with 0:00 am peak has a thicker tail compared with the other one with 10:30 pm peak.
\end{itemize}
Clearly, these two patterns reflect two different types of students: students who stay up later and students who don't stay up late. According to the regulations of the university where we carry out our experiments, all lights in dormitory areas will be turned off at 10:30 to urge students to sleep early, which is consistent with our experimental results, verifying the effectiveness of our proposed method.
Moreover, it can be seen in Table \ref{student number} that from the freshmen to the juniors, the proportion of students who do not stay up late gradually increased, eventually surpassing the proportion of students staying up late. The possible reason is that they are willing to live a more regular life because of increased academic pressure and forthcoming job-hunting.
\begin{figure*}[]
	\subfigure[Freshman]{
		\begin{minipage}{0.3\linewidth}
			\centering
			\includegraphics[width=2in]{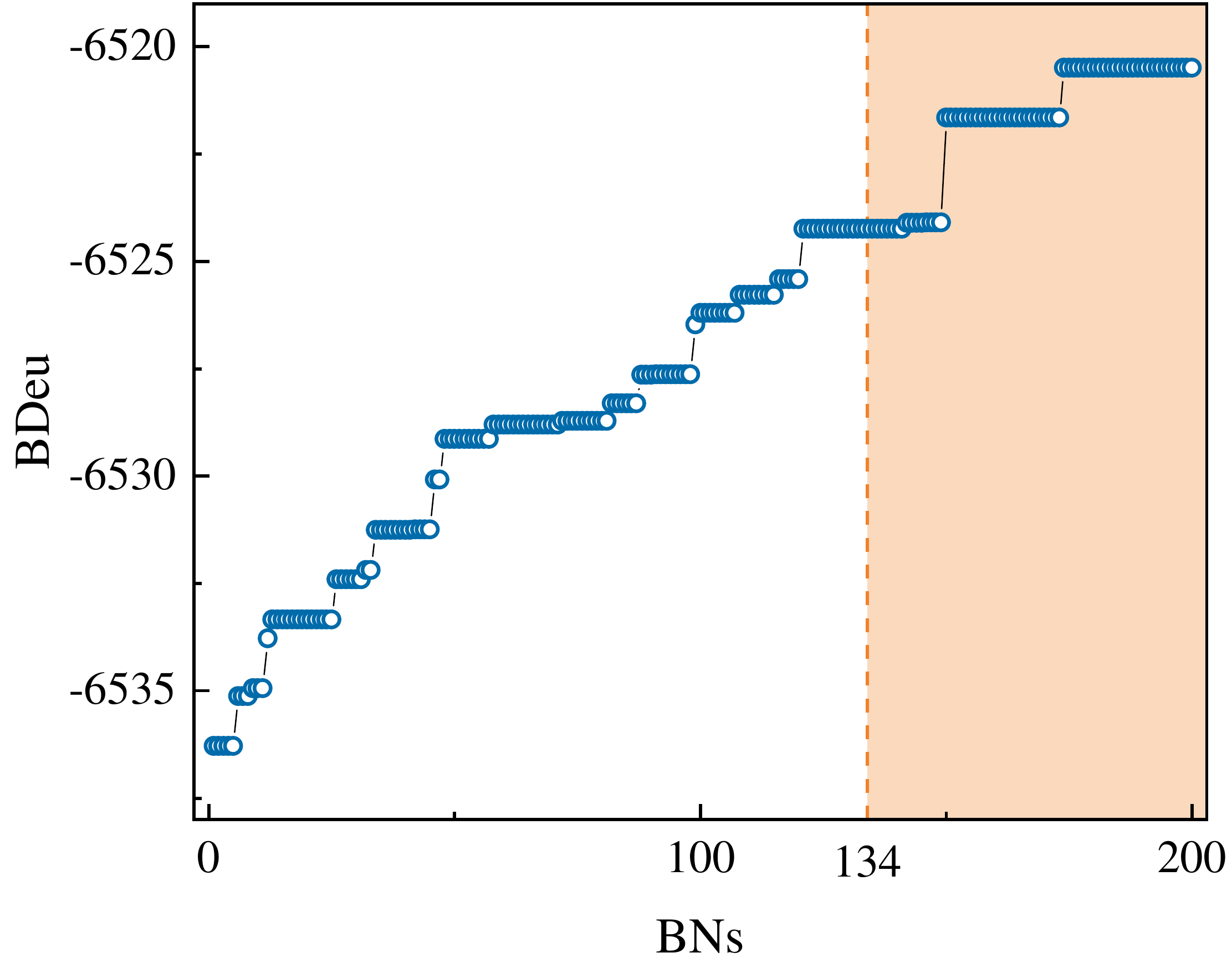}
        \label{example1}
		\end{minipage}
	}
	\subfigure[Sophomore]{
		\begin{minipage}{0.3\linewidth}
			\centering
			\includegraphics[width=2in]{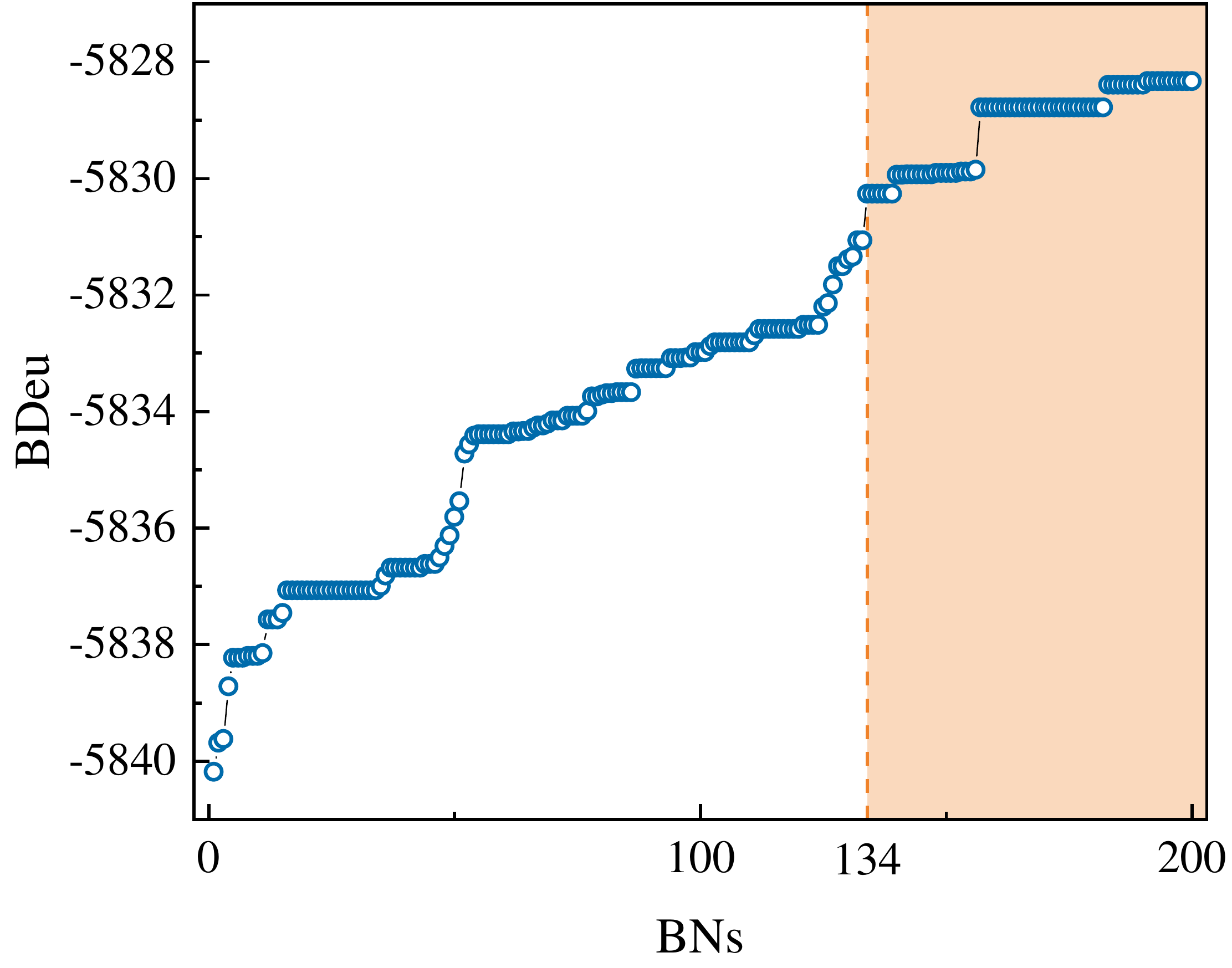}
        \label{example2}
		\end{minipage}
	}	\subfigure[Junior]{
		\begin{minipage}{0.3\linewidth}
			\centering
			\includegraphics[width=2in]{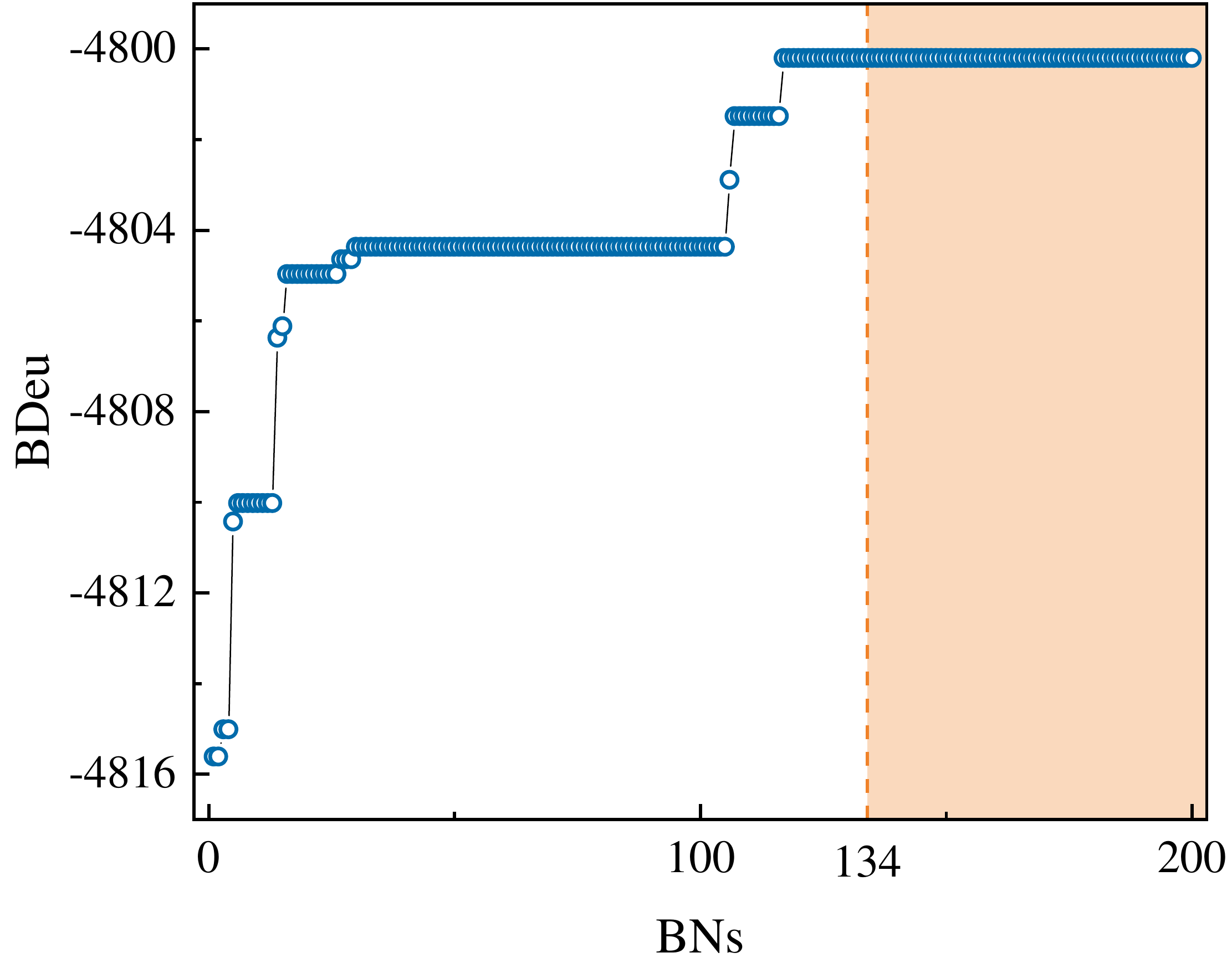}
        \label{example2}
		\end{minipage}
	}	
	\caption{Scores of 200 BNs from restart strategy. We select one-third of the BNs with the high score (shown in the yellow area) for the null model and the consensus network.}
	\label{200scores}
\end{figure*}


\subsection{Student Profiling}
In this section, besides sleep status ($S$) defined above through the mixture probabilistic model, we profile students from other five aspects in eight dimensions, including interest ($R$, $T$ and $A$), orderliness ($Br$ and $Ba$), finance($F$), Academic performance ($Ac$) and Gender ($G$). As shown in Table \ref{variable}, we sort students according to specific indicators and classify students with 50\% as a threshold, resulting in that all variables are binary. The details are introduced in this section.

The interest of students is profiled from three aspects: reading habits $R$, Interest access time $T$, and internet preferences $A$.
First, for reading habit$R$, we divide students into two categories: those who like to read and those who don't, based on the number of books borrowed during the experiment. Second, for Interest access time $T$, we divide students into those who like to surf on the Internet and those who don't, based on their average Interest access time. Finally, for app preference status $A$, apps in our dataset can be classified into two types: games app and video app according to their functions, and compare the time that users spend on both apps to identify their $A$.

The orderliness of students is profiled from two aspects: breakfast orderliness $Br$ and bath orderliness $Ba$. For orderliness profiling, the indicator of $Br$ is the number of taking breakfasts during the experiment and the indicator of $Ba$ is the variance of bath date interval during the experiment. Students are classified into two categories according to the corresponding indicators, with a threshold of 50\%.

In addition to interest and orderliness, we also profile students from academic performance, financial situation, and gender perspectives. For academic performance, students are divided into two categories based on their Grade-Point Average (GPA). For the financial situation, we divided students into two categories according to their daily spending during the experiment. For gender, students are divided into males and females.

\subsection{Bayesian Network Analysis}
\subsubsection{Bayesian Network}

As mentioned above, our dataset includes three types of students: freshmen, sophomores, and juniors. To capture the rules between different groups, we first build BNs for these three groups separately. In this research, we use a hill-climbing strategy to build BNs through optimizing the BDeu score.

\begin{figure*}
	\subfigure[Freshman]{
		\begin{minipage}{0.3\linewidth}
			\centering
			\includegraphics[width=2.1in]{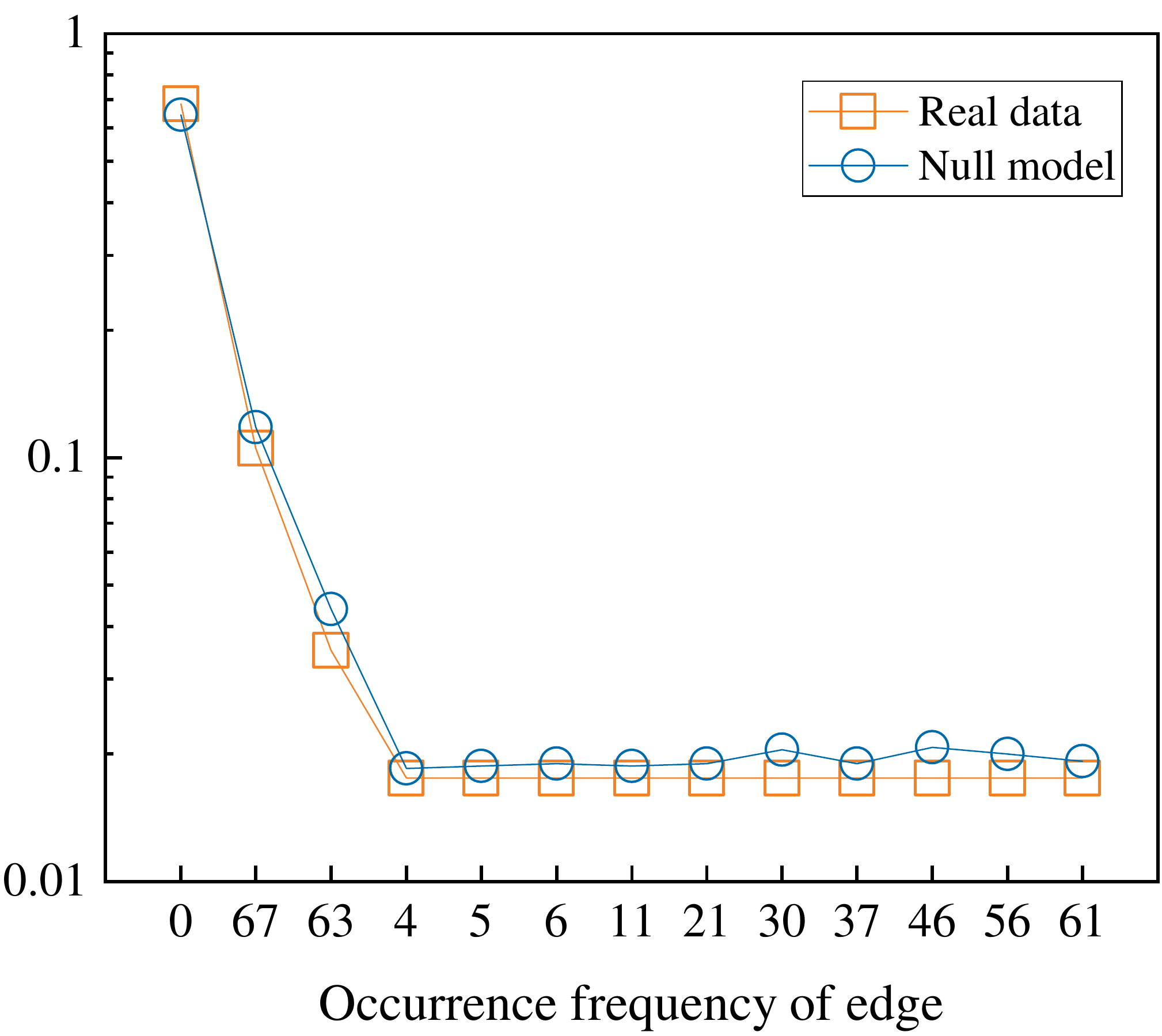}
        \label{n15}
		\end{minipage}
	}
	\subfigure[Sophomore]{
		\begin{minipage}{0.3\linewidth}
			\centering
			\includegraphics[width=2.1in]{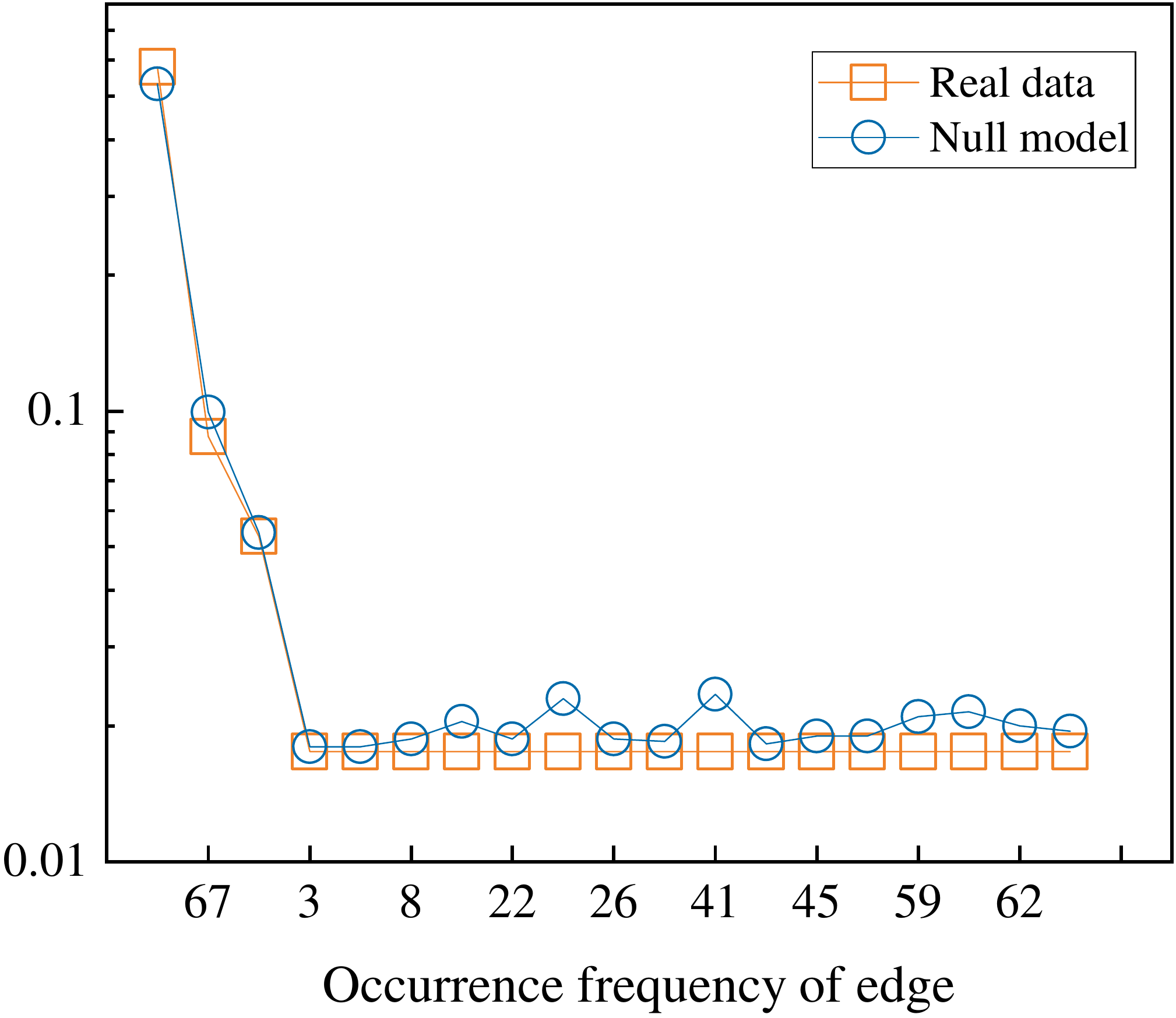}
        \label{n16}
		\end{minipage}
	}	\subfigure[Junior]{
		\begin{minipage}{0.3\linewidth}
			\centering
			\includegraphics[width=2.1in]{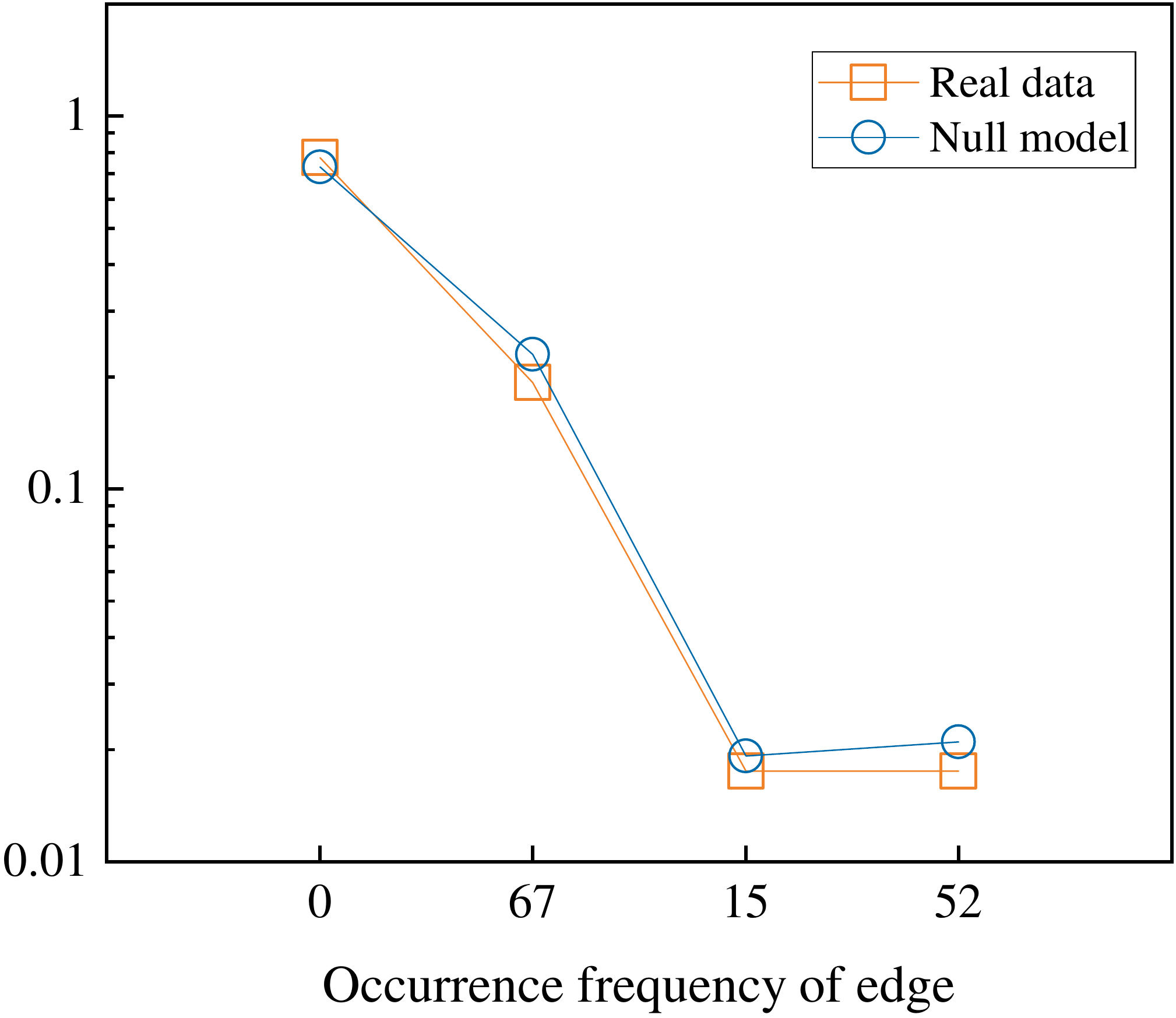}
        \label{n17}
		\end{minipage}
	}	
	\caption{Null model for each group.}
	\label{null}
\end{figure*}

\begin{enumerate}
\item First, for each group, we learn 200 BNs through a hill-climbing strategy with restarts strategy, and the results are shown in Figure \ref{200scores}. We learn 200 BNs and 300 BNs, respectively, and find that the score of high-quality BNs in these two cases are equal.
\item Second, we select one-third of the BNs with the high score for the null model to build the consensus network \cite{agrahari2018applications}. According to the frequency of edge occurring in these BNs, we design null models with bootstrap sampling to filter the invalid edge (shown in Figure \ref{null}) \cite{mcgeachie2014cgbayesnets}. We determine the threshold by keeping the occurring frequency above the mean value plus two times of standard deviation of the random case and in each group, the value is 25, 27, and 25, respectively.
\item Third, we reconstruct the consensus BNs by those edges that occur more frequently than thresholds.
\end{enumerate}

\begin{figure*}
	\subfigure[Freshman]{
		\begin{minipage}{0.22\linewidth}
			\centering
			\includegraphics[width=1.5in]{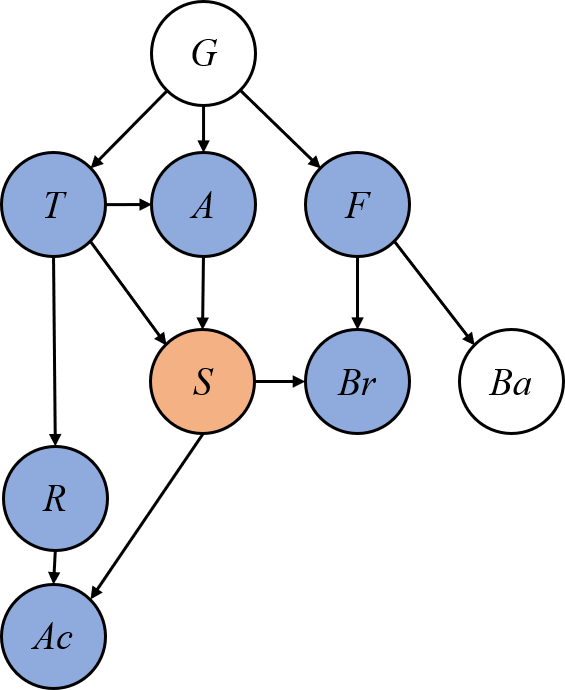}
        \label{bn15}
		\end{minipage}
	}
	\subfigure[Sophomore]{
		\begin{minipage}{0.22\linewidth}
			\centering
			\includegraphics[width=1.3in]{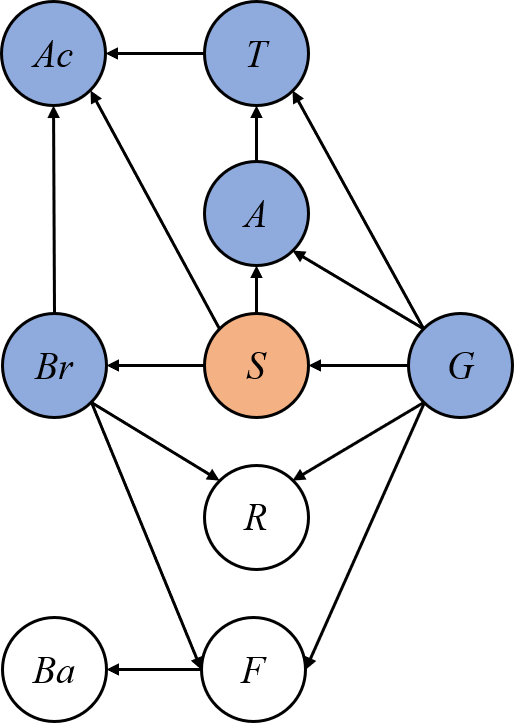}
        \label{bn16}
		\end{minipage}
	}	\subfigure[Junior]{
		\begin{minipage}{0.22\linewidth}
			\centering
			\includegraphics[width=1.5in]{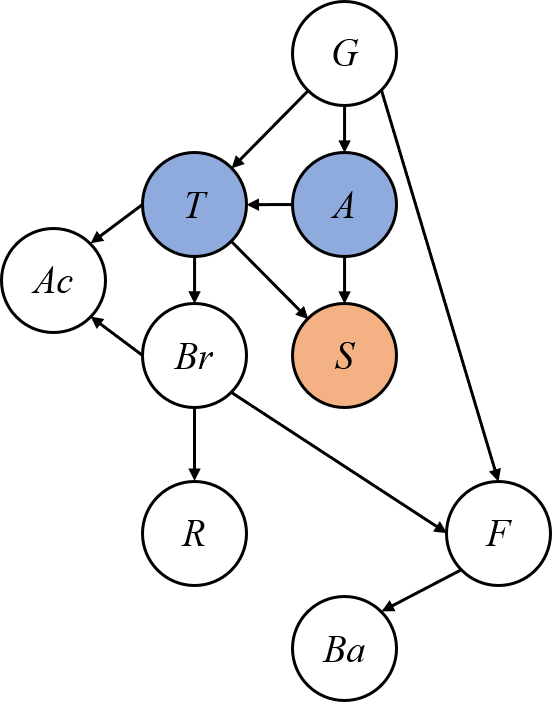}
        \label{bn17}
		\end{minipage}
	}	
	\subfigure[Total]{
		\begin{minipage}{0.22\linewidth}
			\centering
			\includegraphics[width=1.1in]{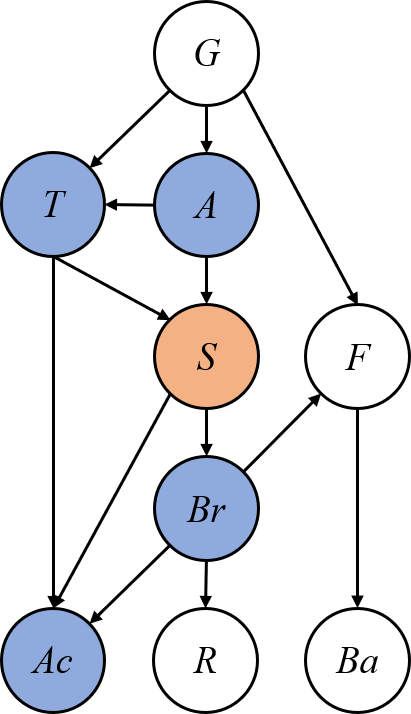}
        \label{bnall}
		\end{minipage}
	}	
	\caption{Consensus BN for each group. The orange node is sleep status and its Markov Blanket is represented by blue node.}
	\label{BNs}
\end{figure*}

All BNs are shown in Figure \ref{BNs}. Moreover, we synthesize a total network based on three consensus-BNs according to the occurring frequency of each edge shown in Figure \ref{bnall}. For the same edge but different directions, we choose the edge with higher occurrence frequency in high score BNs. For example, in three consensus-BNs, the edge from $T$ to $A$ and the edge from $A$ to $T$ have the same occurrence frequency. Because the edge from $A$ to $T$ has a higher occurrence frequency in BNs with a high score, we keep this edge in total BNs.

%
%

\begin{figure*}[]
	\subfigure[Freshman]{
		\begin{minipage}{0.22\linewidth}
			\centering
			\includegraphics[width=1.6in]{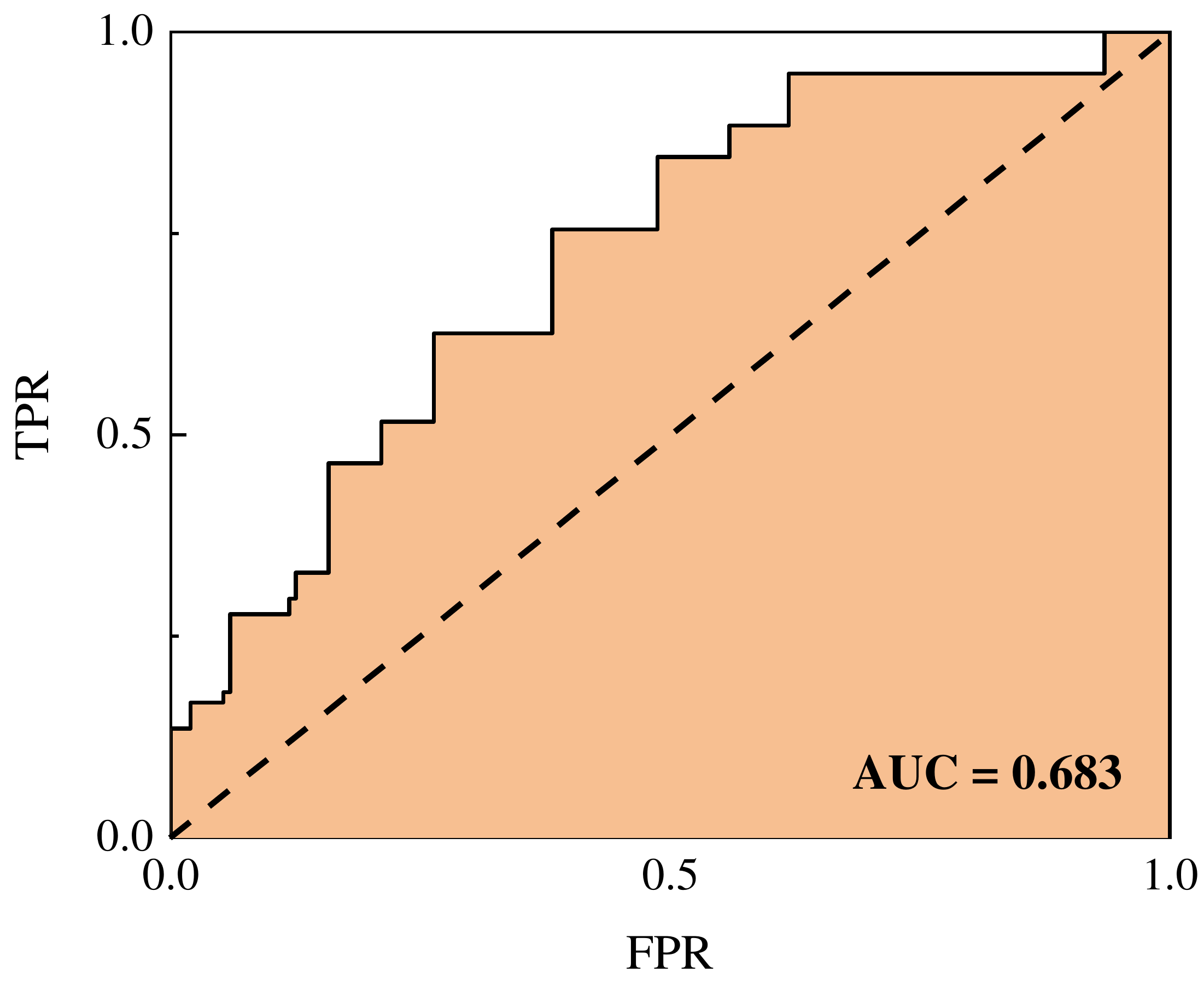}
        \label{example1}
		\end{minipage}
	}
	\subfigure[Sophomore]{
		\begin{minipage}{0.22\linewidth}
			\centering
			\includegraphics[width=1.6in]{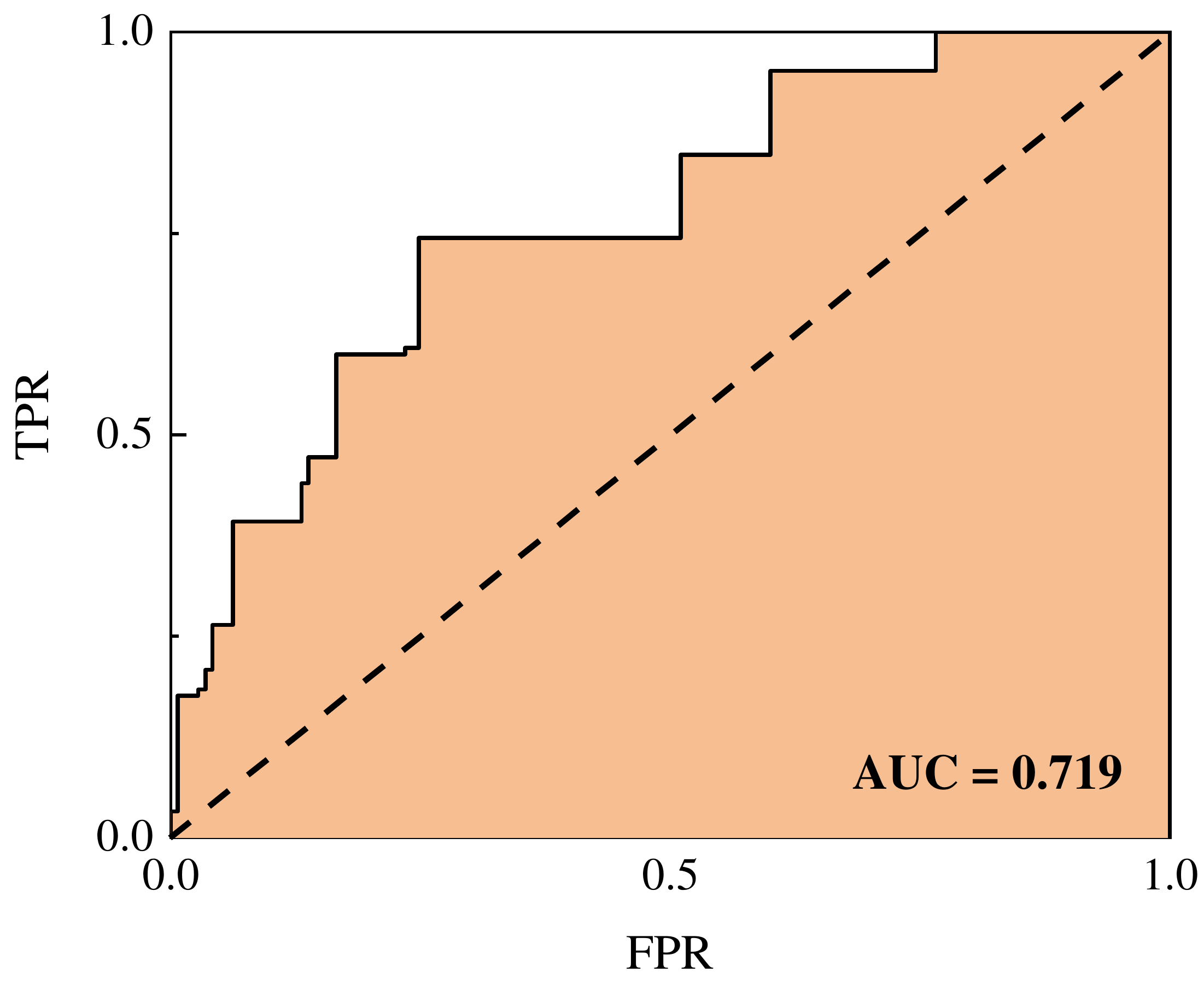}
        \label{example2}
		\end{minipage}
	}	\subfigure[Junior]{
		\begin{minipage}{0.22\linewidth}
			\centering
			\includegraphics[width=1.6in]{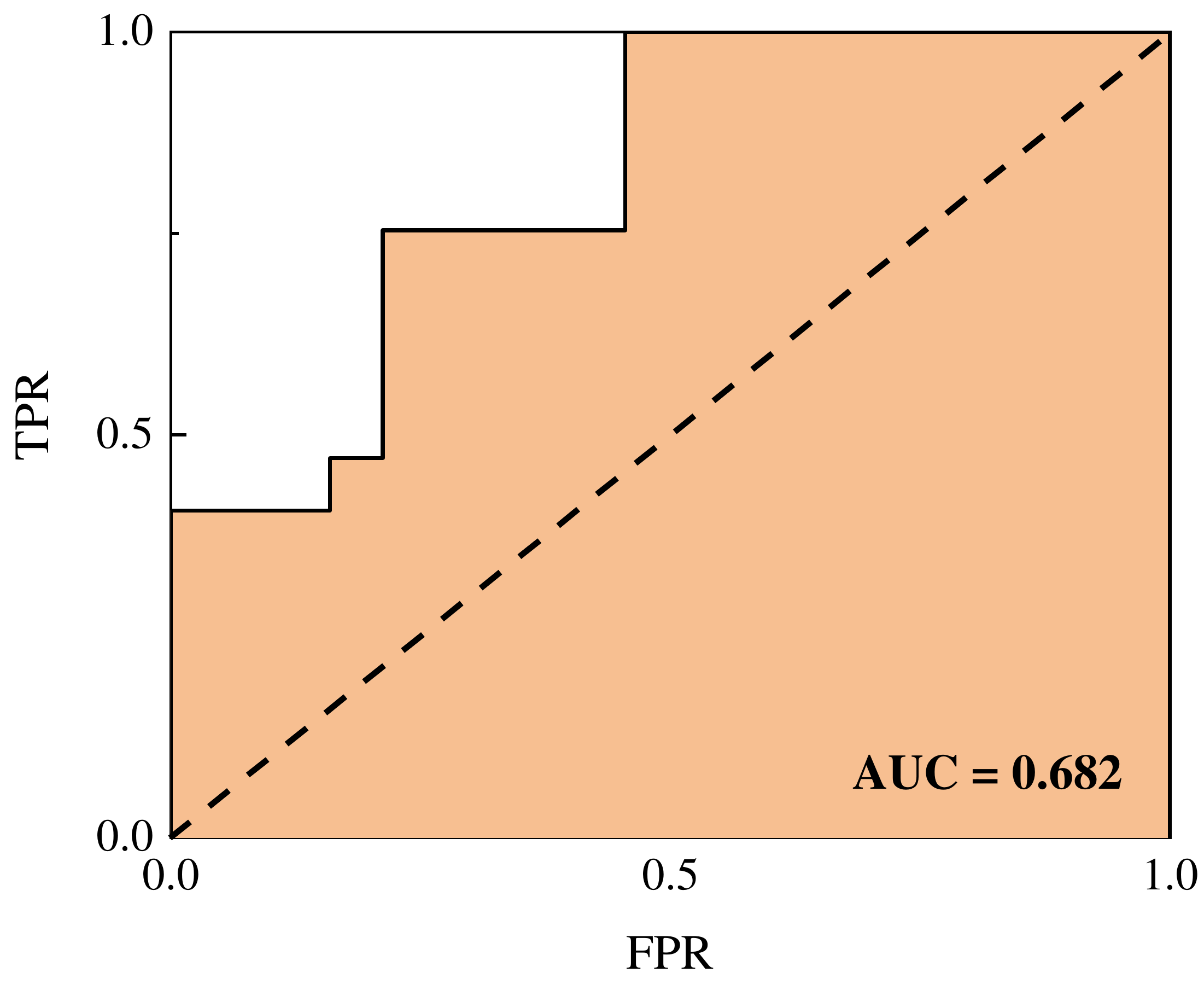}
        \label{example2}
		\end{minipage}
	}	
\subfigure[Total]{
		\begin{minipage}{0.22\linewidth}
			\centering
			\includegraphics[width=1.6in]{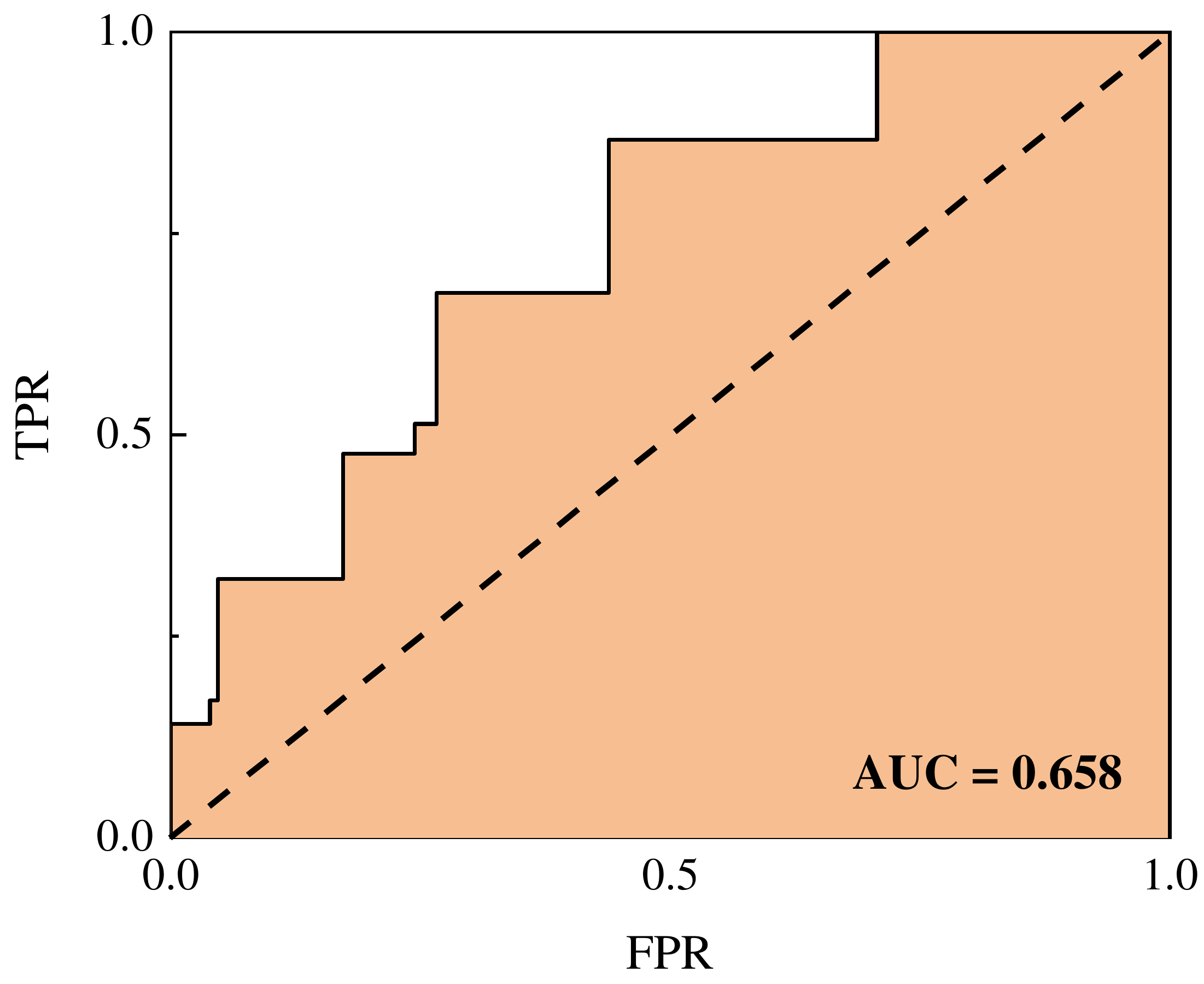}
        \label{example2}
		\end{minipage}
	}	
	\caption{Prediction performance of each BNs. We use ROC to measure the prediction performance. AUC is indicated by yellow and diagonal dotted lines indicate 0.5 in random cases.}
	\label{roc}
\end{figure*}

\subsubsection{Structure Analysis}

\begin{table}[htp]
  \centering
  \caption{Structure analysis for $S$ in the three BNs.}
    \begin{tabular}{l|llllllll}
    \hline
    & $G$& $R$& $A$& $T$& $Br$& $Ba$    & $F$     & $Ac$  \\
    \hline
    Freshman&      & $\surd$     &{\color{red} $\surd$}       & {\color{red} $\surd$ }     & {\color{cyan} $\surd$ }    &    & $\surd$     &{\color{cyan} $\surd$ }  \\
    \hline
    Sophomore& {\color{red}$\surd$ }    &     & $\surd$      & $\surd$      & {\color{cyan} $\surd$ }   &    &    &{\color{cyan} $\surd$ }  \\
    \hline
    Junior&      &     & {\color{red}$\surd$  }    & {\color{red}$\surd$  }    &    &     &     &   \\
    \hline
    Times in $\pi(S)$ & 1     &     & 2   & 2    &    &     &     &   \\
    \hline
    Times in $MB(S)$ & 1     &1     & 3    & 3    &2    &     &  1   & 2  \\
    \hline
    Times in $ch(S)$ &      &     &     &     &2    &     &     & 2  \\
    \hline
    \end{tabular}%
    		\begin{tablenotes}
			\item[]$\surd$: The variable is in $MB(S)$.
            \item[]{\color{red}$\surd$}: The variable is in $\pi(S)$.
            \item[]{\color{cyan}$\surd$}: The variable is in $ch(S)$.
		\end{tablenotes}
  \label{parentsnode}%
\end{table}%

The three networks shown in Figure \ref{BNs} represent three types of students: freshmen, sophomores, and juniors. It is obvious that the behaviors of different groups are different. According to the Markov property of the Bayesian network, the parents node $\pi(X_i)$ are the most related to the output of the targeted node $X_i$. In this case, we summary those BNs in Figure \ref{BNs} and show the results in Table \ref{parentsnode}.

First, both $A$ and $T$ are frequent in $\pi(S)$, representing that a strong connection exists between sleep habit and surfing habit. Specifically, from the perspective of App preference, the probabilities of staying up late for people who love watching videos are 0.75, 0.63, and 0.70, in freshmen, sophomores, and juniors, respectively.
For people who love playing games, the probabilities are 0.5, 0.39, and 0.47. Video lovers prefer to stay up late than game fans. Then, we summarize the most popular games among college students, and the top three are Glory of Kings, PUBG (Playerunknown's Battlegrounds), and LOL (League of Legends), which account for more than 80 percent (This result is consistent with the findings of previous studies that people who primarily engage in group play have superior adherence to people who primarily play alone \cite{kaos2019social}).
A common feature of these games is that they are all web-based battle games. Compared to previous single-player games, these games bring all players together to collaborate or compete with each other through an online platform.
Players play with real people instead of computers, which is closer to social behavior.
In this case, if the user's enemies or teammates are sleeping, then this game naturally loses fun, which explains why game fans don't like staying up late. This phenomenon can also be analyzed from another perspective.
According to the theory mentioned in \cite{ducheneaut2006alone}, 'spectator experience' and a sense of social presence are very important for players of online games, which can also explain the finding of our experiment.
In contrast, watching a video is a behavior that a student can do well on their own, even in the middle of the night. Note that the above results show that people who love to watch videos are more likely to stay up late rather than that videos are to keep people stay up late compared with games, due to the differences in the number of each type of students.

Does the seemingly rational behavior of game lovers contradict the popular view of game addiction? We analyze the relation between app preference $A$ and total surfing length $T$ due to that $A$ appears frequently in the parent node of $T$. As mentioned before, according to the total length of surfing time, we divide students into long-time internet users and short-time internet users. The probabilities of game fans surfing the Internet for a long time are 0.57, 0.60, and 0.57 in three groups, which is higher than 0.25, 0.22, and 0.22 for video lovers. This is in line with the current popular view of game addiction. At the same time, the above mentioned can also explain why the short-time internet user prefers to stay up late. (Note that this finding does not conflict with the conclusions in \cite{orzech2016digital} that a longer duration of digital media use was associated with reduced total sleep time and later bedtime because they focus on the digital media two hours before bedtime instead of all day).

Second, for the child nodes of $S$, the variables of breakfast status $Br$ and academic performance $Ac$ appear frequently according to Table \ref{parentsnode}. In the matter of having breakfast, the probability of students staying up late is 0.2 lower than normal students, which is consistent with common sense that students who like to stay up late usually get up late to miss breakfast.
For the academic performance, students who go to bed early are more likely to achieve a good grade, which is consistent with previous findings, both in biology \cite{kelley2015synchronizing} and cognitive science \cite{althoff2017harnessing}.

\subsubsection{Inference}

The Markov property of the Bayesian network implies that conditioned on the Markov blanket of a node (shown in Figure \ref{BNs}), the probability distribution of the node is independent of the rest of the network shown as the following equation:

\begin{equation}
\begin{aligned}
X\perp\{U-MB(X)-\{X\}\}|MB(X)
\end{aligned}
\label{mb}
\end{equation}
where $U$ is the set of all random variables in BNs and $MB(X)$ represents the Markov blanket of variable $X$, which is a set of nodes that consists of the parents of the node, the children of the node, and any other parents of the children of that node.

In our research, we design an experiment of predictive inference to estimate the predictability of sleep status. In other words, we want to test whether the sleep state $S$ can be predicted rather than pursue prediction performance. So, we use MLE to fit our datasets and use Area Under The Curve (AUC) to evaluate the prediction performance. For AUC, if the AUC is bigger than 0.5, representing that the prediction result is better than the random model. In other words, the $S$ is predictive. From Figure \ref{roc}, the AUC value for each group is 0.683, 0.719 and 0.682, which reflects the predictability of sleep pattern.

\section{Discussion and Conclusion}

In this paper, we design a Possion-based probabilistic mixture model to identify students who are used to stay up late based on the Internet access patterns. We apply this model to a real-world dataset and classify students into two groups: students who are used to stay up late and students who sleep on time. We profile students from five aspects in eight dimensions, including interest (reading status, app preference status, and surfing length status), orderliness (breakfast orderliness status and bath orderliness status), finance (financial status), academic performance (academic performance status) and gender. Then we build Bayesian networks to explore the relationship between these characteristics and sleeping habits and find that surfing habits have a big impact on sleep habits. Finally, we test the predictability of sleeping habits based on campus behavior features.

The assumption of our experiment that students will access the Internet through mobile phones or computers before bedtime is reasonable because the Internet pervades every aspect of our lives, including entertainment, study, social contact, and so on.
However, the underlying assumptions made in this study raise a couple of limitations. First, we identify students staying up late based on a hypothesis that
students use mobile phones or computers to access the Internet before going to bed. So, it is hard to detect the people who don't have this habit. Second, more data of life details need to be collected for drawing a valid and solid conclusion.

There are multiple avenues for future work. First, we intend to expand our dataset and investigate this issue from more aspects. Second, we only check if the sleep status can be predicted rather than pursuing a precise prediction. Next, we plan to design a prediction model with a good performance and integrate this model into the modern educational management system and apply real-time data to detect the sleep status of students.

\bibliographystyle{abbrv}
\bibliography{sigkddExp-0917}

\end{document}